\documentclass[runningheads,a4paper]{llncs}

\usepackage{amssymb}
\usepackage{amsmath}
\usepackage{graphicx}
\usepackage{color}
\usepackage{booktabs}
\usepackage{subfigure}

\usepackage{url}

\begin{document}

\mainmatter 

\title{Semi-Supervised Deep Learning for Fully Convolutional Networks}

\author{Christoph Baur\inst{1} \and Shadi Albarqouni\inst{1}\thanks{C. Baur and S. Albarqouni contributed equally to this work.} \and Nassir Navab\inst{1,2}}

\authorrunning{Baur et al.}

\institute{Computer Aided Medical Procedures (CAMP), Technical University of Munich, Munich, Germany\\
\and Whiting School of Engineering, Johns Hopkins University, Baltimore, United States}

\maketitle

\begin{abstract}
Deep learning usually requires large amounts of labeled training data, but annotating data is costly and tedious. The framework of semi-supervised learning provides the means to use both labeled data and arbitrary amounts of unlabeled data for training. Recently, semi-supervised deep learning has been intensively studied for standard CNN architectures. However, Fully Convolutional Networks (FCNs) set the state-of-the-art for many image segmentation tasks. To the best of our knowledge, there is no existing semi-supervised learning method for such FCNs yet. We lift the concept of auxiliary manifold embedding for semi-supervised learning to FCNs with the help of \emph{Random Feature Embedding}. In our experiments on the challenging task of MS Lesion Segmentation, we leverage the proposed framework for the purpose of domain adaptation and report substantial improvements over the baseline model.

\end{abstract}

\section{Introduction}

In order to train deep neural networks, usually huge amounts of labeled data are neccessary. In the medical field, however, labeled data is scarce as manual annotation is time-consuming and tedious. At the same time, when training models using a limited amount of labeled data, there is no guarantee that these models will generalize well on unseen data that is distributed slightly different. A prominent example in this context is Multiple Sclerosis lesion segmentation in MR images, which suffers from both a lack of ground-truth and distribution-shift across images from different devices~\cite{garcia2013review}. However, vast amounts of unlabeled data can often be provided comparably easy. Semi-supervised learning provides the means to leverage both a limited amount of labeled data and arbitrary amounts of unlabeled data for training deep networks~\cite{weston2012deep}. In recent years, various frameworks for semi-supervised deep learning have been proposed: In 2012, Weston et al.~\cite{weston2012deep} presented a framework for artificial neural networks and shallow CNNs based on auxiliary manifold embedding. Using an additional embedding loss function attached to arbitrary hidden layers and graph adjacency among input samples, they forced the feature representations of neighbouring labeled and unlabeled samples to become more similar, leading to improved generalization. In 2013, Lee et al.~\cite{lee2013pseudo} also reported improved generalization when fine-tuning a model from predictions on unlabeled data using an entropy regularization loss. More recently, in 2015, Rasmus et al.~\cite{rasmus2015semi} introduced the ladder network architecture for semi-supervised deep learning. And lately, Yang et al.~\cite{yang2016revisiting} presented a framework also based on graph embeddings, with both transductive and inductive variants for shallow neural networks.

Even though these methods show promising results, all of them are tailored to classic CNN architectures and are often only examined on small scale computer vision datasets. In challenging problems such as biomedical image segmentation, Fully Convolutional Networks (FCNs)~\cite{long2015fully} are preferable as they are efficient and show the ability to learn context~\cite{milletari2016v,kamnitsas2016efficient}. As far as we know, there is no existing semi-supervised learning method for such FCNs yet.
In this paper, we lift the concept of auxiliary manifold embedding to FCNs with a, to the best of our knowledge, novel strategy called ``Random Feature Embedding''. Subsequently, we successfully perform semi-supervised fine-tuning of FCNs for domain adaptation with our proposed embedding technique on the challenging task of Multiple Sclerosis (MS) lesion segmentation.

\section{Methodology} 
\label{sec:methodology}

\begin{figure}[t]
\centering
\includegraphics[width=\textwidth]{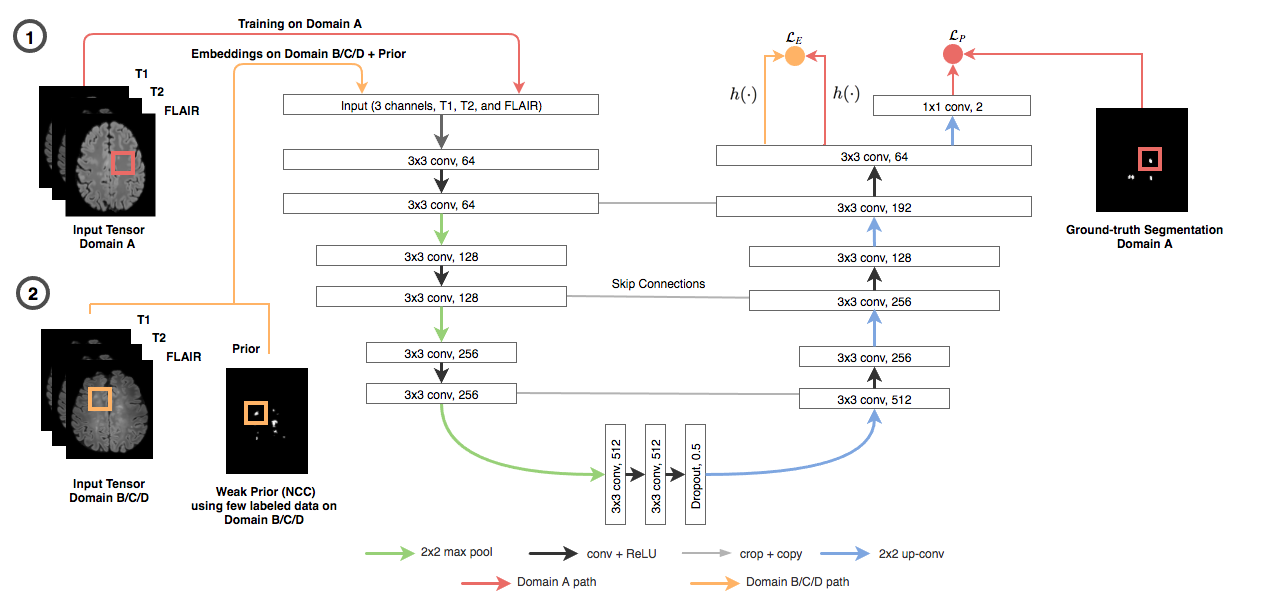}
\caption{Illustration of the semi-supervised deep learning framework.}
\label{fig:frameworkDepiction}
\end{figure}

Semi-supervised learning is based on the concept of training a model, $f(\cdot)$, using both labeled and unlabeled data $\mathbf{X_L}$  and $\mathbf{X_U}$, respectively. In our framework for FCNs, training data has the form of  $\mathcal{\mathbf{D}}=\{\mathbf{X}, \mathbf{Y}\}$, where $\mathbf{X} = \{\mathbf{X_L} \cup \mathbf{X_U}\} =\{\mathbf{x_1},..., \mathbf{x_{N_L}}, \mathbf{x_{N_{L+1}}},..., \mathbf{x_{N_{L+U}}}\} \in \mathbb{R}^{H\times W \times D \times N_{L+U}}$ are $D$-channel images, and $\mathbf{Y} = \{\mathbf{y_1},...,\mathbf{y_{N_L}}\} \in \mathbb{R}^{H\times W \times 1 \times N_L}$ are the corresponding label maps which are only available for the labeled data. Since we deal with FCNs, both images and label maps have the same dimensions, $H \times W$, allowing a distinct label per pixel. 

\subsection{Auxiliary Manifold Embedding}
\label{sub:embedding}
In our framework, to model $f(\cdot)$, we employ a modified version of the U-Net architecture~\cite{ronneberger2015u} that processes images of arbitrary sizes and outputs a label map of the same size as the input (Fig. \ref{fig:frameworkDepiction}). We train the network to minimize the primary objective $\mathcal{L}_{P}$, i.e. the Dice-Loss~\cite{milletari2016v}, for our segmentation task (see Fig. \ref{fig:frameworkDepiction}) from labeled data only (Fig. \ref{fig:frameworkDepiction}, step 1). Simultaneously, to leverage the unlabeled data, we employ an auxiliary manifold embedding loss $\mathcal{L}_{E}$ on the latent feature representations $h(\cdot)$ of both $\mathbf{X_L}$  and $\mathbf{X_U}$ to minimize the discrepancy between similar inputs in the latent space. Thereby, similarity among $h(\cdot)$ of unlabeled data is given by prior knowledge (Fig. \ref{fig:frameworkDepiction}, step 2). The overall objective function can be written using Lagrangian multipliers as: 
\begin{equation}
\mathcal{L} = \mathcal{L}_{P} + \sum_{l} \lambda_l \cdot \mathcal{L}_{E_l}
\end{equation}
where $\lambda_l$ is the regularization parameter associated with the embedding loss $E_l$ at hidden layer $l$. Typically, this embedding loss function aims at minimizing the distance among latent representations of similar $h^l(\mathbf{x_i})$ and $h^l(\mathbf{x_j})$ of neighboring data samples $\mathbf{x_i}$ and $\mathbf{x_j}$, and otherwise tries to push them apart if their distance is within a margin $m$:
\begin{equation}
\mathcal{L}_{E_l}(\mathbf{X}, \mathbf{A}) = \sum_{i}^{n_E} \sum_{j}^{n_E}\begin{cases}
d(h^l(\mathbf{x_i}), h^l(\mathbf{x_j})), & \text{if $a_{ij} = 1$}\\
\max (0, m - d(h^l(\mathbf{x_i}), h^l(\mathbf{x_j}))), & \text{if $a_{ij} = 0$}
\end{cases},
\end{equation}

Thereby, $\mathbf{A} \in \mathbb{R}^{n_E \times n_E}$ is an adjacency matrix between all embedding samples $n_E$ within a training batch, and $d(\cdot, \cdot) \in \mathbb{R}^1$ is an arbitrary distance metric measuring the distance between two latent representations. Unlike the typical $\ell_2$-norm distance employed in~\cite{weston2012deep}, we opt for the angular cosine distance (ACD) for two reasons; first, it is naturally bounded between $[0,1]$, hence limits the searching area for the marginal distance parameter $m$, and second, it shows superior performance on high-dimensional feature representations in deep architectures~\cite{hinton2010rectified}. 

The definition of $\mathbf{A}$ is left to the user and can be arbitrary. A useful notion of adjacency is for instance given by neighbouring frames in a video stream~\cite{weston2012deep}. We either compute the coefficients of $\mathbf{A}$ based on the label maps, or based on results from template matching with NCC, similar to \cite{bermudez2016scalable}. 

\subsection{Random Feature Embedding} 
\label{sub:random_feature_embedding}

\begin{figure}[t]
\centering
\includegraphics[width=0.9\textwidth]{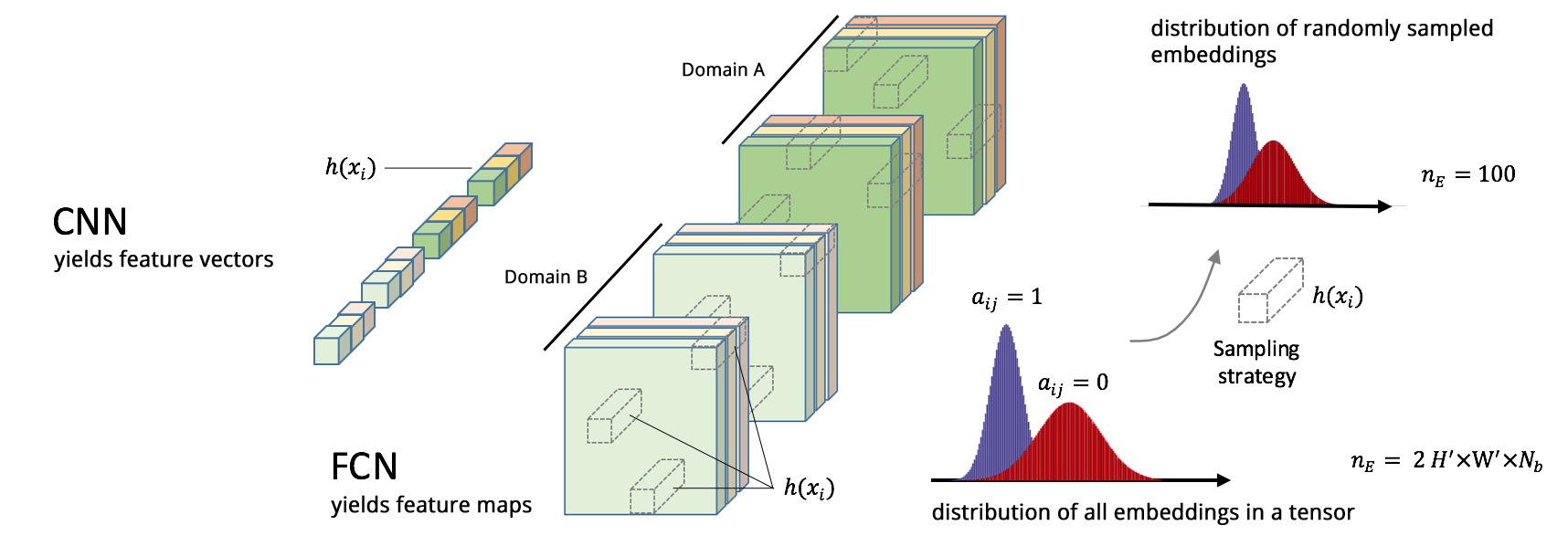}
\caption{For FCNs, we sample embeddings $h(\mathbf{x_i})$ of single pixels from feature maps along the channel dimension. Randomly sampled embeddings should be representative for the entire population.}
\label{fig:randomFeatureEmbeddingTensor}
\end{figure}

In standard CNNs, the embedding loss is commonly attached to the fully connected layers, where a latent feature vector $h(\mathbf{x_i})$ represents a single input image $\mathbf{x_i}$. In the case of FCNs, however, an input of arbitrary size can produce a multi-channel feature map of arbitrary size. Since FCNs make predictions at the pixel level, meaningful embeddings can be obtained per pixel along the channel dimension (Fig. \ref{fig:randomFeatureEmbeddingTensor}). However, sampling and comparing all $h(\cdot)$ of all pixels in large images is computationally infeasible: a $W' \times H' \times D \times N_b$ feature map tensor for a batch of size $N_b$ will yield $n_E = W' \cdot H' \cdot N_b$ and therefore $n_E^2$ comparisons. This quickly becomes intractable to compute. Instead, we suggest to do \emph{Random Feature Embedding} (RFE), where we randomly sample a limited number $n_{E}$ of pixels from the feature maps of the current batch according to some sampling strategy discussed in the next section to limit the number of comparisons. The loss remains valid since we propagate back only the gradients of selected pixels.

\paragraph{Sampling strategy.} 

Ideally, the distribution of randomly sampled embeddings should mimic the one of all embeddings (Fig. \ref{fig:randomFeatureEmbeddingTensor}), while at the same time paying attention to the class distribution such that unwanted bias is not introduced to the model. Therefore, we investigate the following sampling strategies:

\begin{itemize}
	\item \emph{50/50 RFE}: For each class in the prior the same amount of embeddings is randomly extracted to represent $h(\cdot)$ from different classes equally. For unbalanced classes, this might lead to oversampling of one class.
	\item \emph{Distribution-Aware RFE}: Embeddings are sampled from the given training batch according to the ratio of negative and positive classes in the prior to preserve the actual class distribution. When classes are unbalanced and $n_E$ is too small, this might lead to undersampling of one class.
	\item \emph{80/20 RFE}: As a trade-off, embeddings can be randomly sampled from a predefined ratio of 80\% background and 20\% foreground pixels.
\end{itemize}

\section{Experiments and Results}
\label{sec:experiments}

Our experiments on MS lesion segmentation are motivated by the fact that existing automatic segmentation methods often fail to generalize well to MRI data acquired with different devices~\cite{garcia2013review}. In this context, we leverage our semi-supervised learning framework for domain adaptation, i.e. we try to improve generalization of a baseline model by fine-tuning it with unlabeled data from other domains. Therefore, using an optimal prior for the adjacency matrix $\mathbf{A}$, we first assess different sampling strategies, the impact of different numbers of embeddings as well as different distance measures for RFE. In succession, we utilize the most promising sampling strategy and distance measure together with a real prior for domain adaptation.

\paragraph{Dataset.}

\begin{table}[t]
	\centering
	\caption{Overview of our MS lesion data and our training / testing split}
\begin{tabular*}{0.95\textwidth}{@{}lllll@{}}
\toprule
Dataset 				& Domain  	& Patients (Train / Test)	& Resolution					& Scanner								\\ \toprule
MSSEG     			& A 				& 3	/ 2										& 144x512x512\hspace{2mm}  					& 3T Philips Ingenia		\\ \midrule
								& B 				& 3	/ 2										& 144x512x512  					& 1.5T Siemens Aera 		\\ \midrule
								& C 				& 3	/ 2										& 144x512x512  					& 3T Siemens Verio  		\\ \midrule
MSKRI 					&	D					&	3	/ 10									& 300x256x256  					&	3T Philips Achieva 		\\ \midrule
\end{tabular*}
\label{table:dataset}
\end{table}

Our MRI MS Lesion data is a combination of the publicly available MSSEG\footnote{https://portal.fli-iam.irisa.fr/msseg-challenge/overview} training dataset and the non-publicly available MSKRI dataset (c.f. Table \ref{table:dataset}) acquired at the Neuroradiology department of Klinikum Rechts der Isar. For every patient there are coregistered T1, T2 and FLAIR volumes as well as a corresponding ground-truth segmentation volume. These are grouped into domains A, B, C and D based on the respective scanner they have been acquired with. For training and validation, we randomly crop patches around lesions from corresponding T1, T2 and FLAIR axial slices, resulting in 128x128x3 sized input tensors and corresponding binary label maps. Per domain, we thereby approx. crop 6000 patches. These are randomly split into training \& validation sets according to a ratio of 7:3. Actual testing is performed on full slices of the testing volumes.

\paragraph{Implementation}
Our framework is built on top of MatConvNet~\cite{vedaldi15matconvnet}. All models were trained in batches of 12 and the learning rate of the primary objective fixed to 10e-6. For embedding with $\ell_2$, we set $\lambda = 0.01$ and the margin parameter $m = 1000$ (empirically chosen), for embedding with ACD we use $\lambda = 1$ and $m = 1$, such that $h(\cdot)$ from different classes become orthogonal.

\subsection{Baseline Models}
\label{sub:MSBaselineModels}

In order to measure the impact of our auxiliary manifold embedding, we first train so called lower bound and upper bound baseline models in a totally supervised fashion from labeled data only. We train the \textbf{Lower Bound Model} $A_{L}$ from domain A training data for 50 epochs to obtain a model which produces decent segmentations on data from domain A, but does not generalize well to other domains. Further, we train so called \textbf{Upper Bound Models} by taking $A_L$ after 35 epochs and fine-tuning it until epoch 50 using mixed, labeled training data from domain A and $d \in [B,C,D]$. We obtain three different models which should ideally be able to segment MS lesion volumes from domain A \& B, A \& C or A \& D respectively.

\subsection{Semi-Supervised Embedding}
\label{sub:semi_supervised_embedding}

For the purpose of semi-supervised deep learning, we now assume there is labeled data from domain A, and multiple unlabeled or barely labeled MRI data from the other domains $d \in [B,C,D]$. All of the models trained in the following experiments originally build upon epoch 35 of the lower bound model A$_L$. One embedding loss is attached to the second-last convolutional layer of the network (Fig. \ref{fig:frameworkDepiction}). This choice is due to the fact that this particular layer produces feature maps at the same resolution as the input images, thus there is no need to downsample the prior required for embedding and thus no risk involved in losing heavily underrepresented, small lesion pixels (the lesion load within volumes is often less than 1\%).

\subsubsection{Proof of Concept}
\label{ssub:proof_of_concept}

\begin{figure}[t]
\centering
\subfigure[]{
		\includegraphics[width=0.47\textwidth]{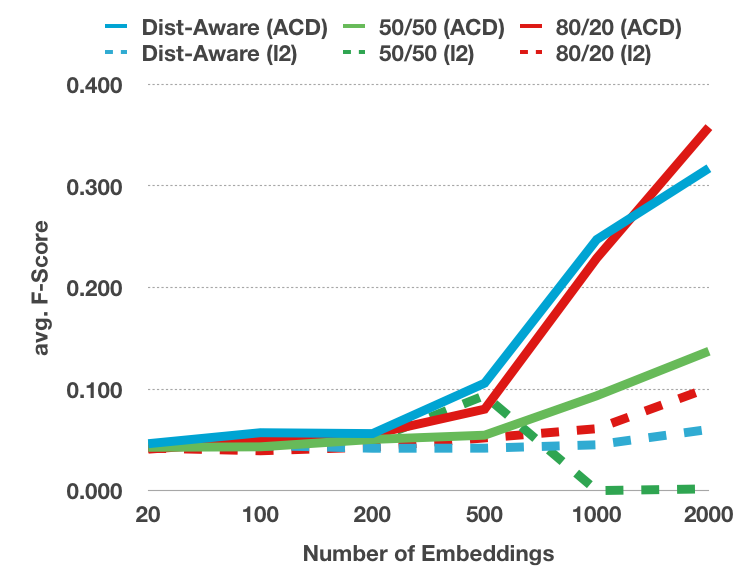}
    \label{fig:proofofconcept}
}
\subfigure[]{
		\includegraphics[width=0.47\textwidth]{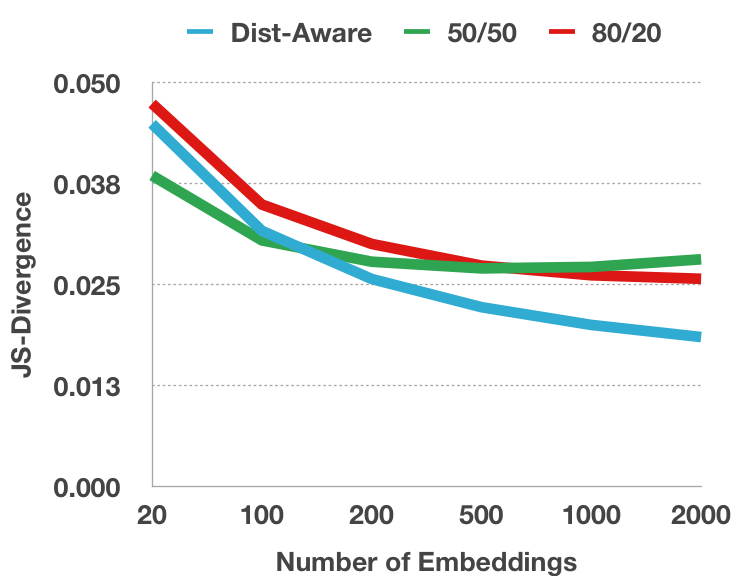}
    \label{fig:jsdiv}
}
\caption{a) Average F-Scores reported on domain B testing data for models trained with different settings and b) the impact of increasing $n_E$ on the Jensen-Shannon divergence between randomly sampled embeddings and the space of all embeddings.}
\end{figure}

We first assume a perfect prior and concentrate on investigating the best choice of sampling strategy, number of embeddings $n_E$ and distance metric. We construct a perfect prior such that $a_{i,j} = 1$ if the labels of $h(x_i)$ and $h(x_j)$ are equal and $a_{i,j} = 0$ otherwise. Based on this prior, we fine-tune models for target domain B using i) 50/50, Distribution-Aware and 80/20 RFE with ii) the $\ell_2$-norm and the ACD as a distance metric and iii) different number of embeddings $n_E \in \{20, 100, 200, 500, 1000, 2000\}$, yielding a total of 36 different models. For fine-tuning in these proof-of-concept experiments, we use only a subset of 200 images from domain A and target domain B each, rather than the full training set. Our results show that, as we ramp up $n_E$, the distribution of randomly sampled embeddings more and more resembles the full distribution (Fig. \ref{fig:jsdiv}), which renders the random sampling generally valid. Moreover, with increasing $n_E$, we notice consistent segmentation improvements on target domain B when using ACD as a distance metric (Fig. \ref{fig:proofofconcept}). In repeated experiments, we always obtain similar results. The improvements are most pronounced with the 80/20 sampling strategy. However, the $\ell_2$-norm performs poorly and seems unstable as the number of embeddings $n_E$ increases. We believe this is because the $\ell_2$-norm penalizes the magnitudes of the vectors being compared, whereas ACD is scale-invariant and only constrains their direction.

\subsubsection{Real Prior}
\label{subsub:real_prior}

\begin{figure}[t]
\centering
\subfigure[]{
		\includegraphics[width=0.4\textwidth]{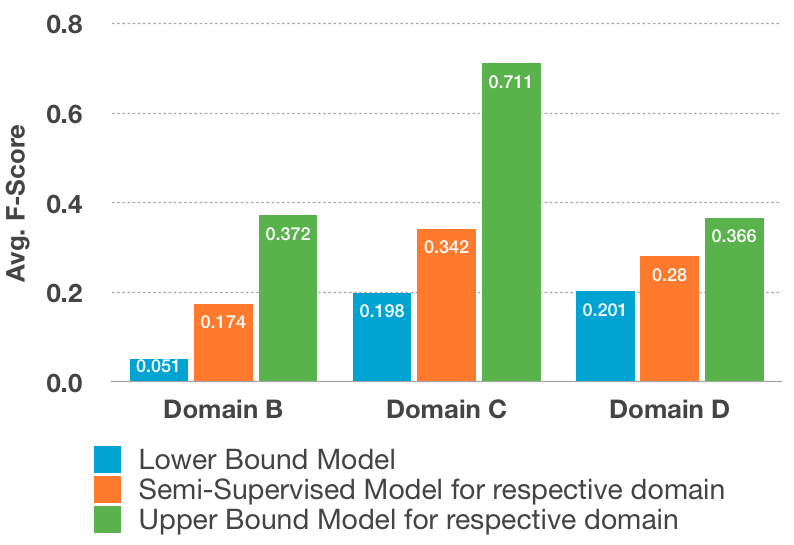}
    \label{fig:nccimpact}
}
\subfigure[]{
		\includegraphics[width=0.55\textwidth]{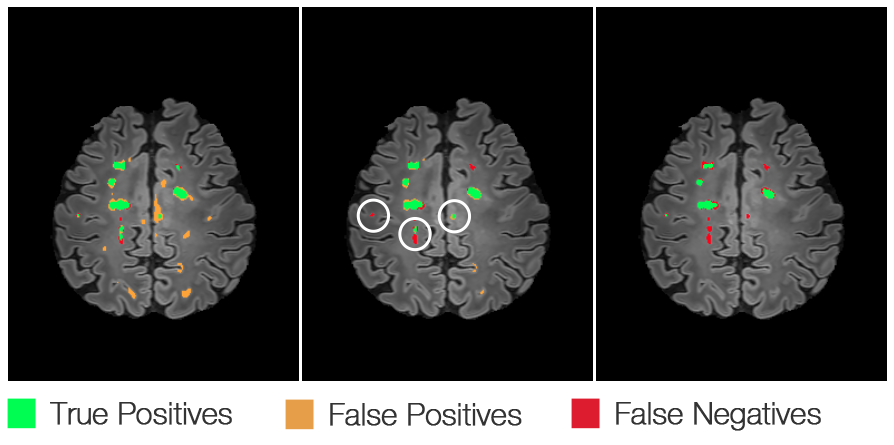}
    \label{fig:examples}
}
\caption{a) Comparing the lower bound model, the semi-supervised models fine-tuned with NCC \& ACD, and the upper bound models on the respective target domain; b) The semi-supervised model for domain C (middle) produces much fewer FP than the lower bound model (left image) and is only slightly inferior to the respective upper bound model (right image).}
\end{figure}

Motivated by previous results, we now train models with a real, noisy prior using ACD and 80/20 RFE. For a target domain $d \in [B,C,D]$, we obtain the prior by selecting the first labeled FLAIR training volume $V_1$ of $d$ and randomly extracting $5 \times 5 \times 5$ voxel sized 3D templates around MS lesions. Using 30 different, randomly sampled 3D templates, we perform NCC on the remaining volumes $V_2$ and $V_3$ of the current domain $d$. For thresholding the template matching output, the same matching is applied to $V_1$ itself. Using the geometric mean of the responses and the ground-truth labels, a threshold is chosen which maximizes the Dice-Coefficient for $V_1$.
Using this noisy prior, we fine-tune models for domain B, C and D using approx. 4000 training patches from volumes $V_2$ and $V_3$ and all labeled training patches from domain A. We set $n_E = 100$ to obtain an overall number of embeddings similar to the proof of concept experiment ($n_E = 2000$), but with lower computational cost. The models show consistent improvements over the lower bound model A$_L$ for all target domains (see. Fig. \ref{fig:nccimpact}). Visual inspection (Fig. \ref{fig:examples}) reveals that the semi-supervised embedding seems to dramatically reduce the number of false positives (FP). Interestingly, it detects some lesions (encircled) where the upper bound model fails, but it is not able to spot very small lesions. This is probably due to the fact that embeddings of smaller lesions are more unlikely to be sampled.

\section{Discussion and Conclusion}
\label{sec:conclusion}

In summary, we presented the concept of auxiliary manifold embedding and successfully integrated it into a semi-supervised deep learning framework for FCNs. At the heart of this framework is \emph{Random Feature Embedding}, a simple method for sampling feature representations which serve as a statistic for non-linear embedding. Our experiments on MS lesion segmentation revealed that the method can improve generalization capabilities of existing models when using ACD as a distance metric. Yet, there is a lot of room for follow-up investigations. For instance, in future work, the effect of attaching multiple embedding objectives at different layers of an FCN could be investigated. In general, the proposed method should be applied to other problems as well to reveal its full potential.

\section{Acknowledgement}
\label{sec:ack}

We thank our clinical partners, in particular Dr. med. Paul Eichinger and Dr. med. Benedikt Wiestler, from the Neuroradiology department of Klinikum Rechts der Isar for providing us with their MRI MS Lesion dataset.

\bibliography{paper}
\bibliographystyle{plain}

\end{document}